\documentclass{article}

\usepackage{microtype}
\usepackage{graphicx}
\usepackage{subfigure}
\usepackage{booktabs} 

\usepackage{hyperref}
\usepackage{enumitem}
\usepackage[table]{xcolor}

\usepackage[accepted]{icml2024}
\usepackage{xcolor}

\usepackage{amsmath}
\usepackage{amssymb}
\usepackage{mathtools}
\usepackage{amsthm}
\usepackage{bbm}

\usepackage[capitalize,noabbrev]{cleveref}

\theoremstyle{plain}

\theoremstyle{definition}

\theoremstyle{remark}

\usepackage[textsize=tiny]{todonotes}

\icmltitlerunning{Exploring Training on Heterogeneous Data with Mixture of Low-rank Adapters}

\begin{document}

\twocolumn[
\icmltitle{Exploring Training on Heterogeneous Data with Mixture of Low-rank Adapters}




\begin{icmlauthorlist}
\icmlauthor{Yuhang Zhou}{sjtu,ai}
\icmlauthor{Zihua Zhao}{sjtu,ai}
\icmlauthor{Siyuan Du}{fudan,ai}
\icmlauthor{Haolin Li}{fudan,ai}
\icmlauthor{Jiangchao Yao}{sjtu,ai}
\icmlauthor{Ya Zhang}{sjtu,ai}
\icmlauthor{Yanfeng Wang}{sjtu,ai}
\end{icmlauthorlist}

\icmlaffiliation{sjtu}{Cooperative Medianet Innovation Center, Shanghai Jiao Tong University}
\icmlaffiliation{ai}{Shanghai Artificial Intelligence Laboratory}
\icmlaffiliation{fudan}{Fudan University}

\icmlcorrespondingauthor{Jiangchao Yao}{Sunarker@sjtu.edu.cn}
\icmlcorrespondingauthor{Yanfeng Wang}{wangyanfeng622@sjtu.edu.cn}


\icmlkeywords{Machine Learning, ICML}

\vskip 0.3in
]



\printAffiliationsAndNotice{}  

\begin{abstract}
Training a unified model to take multiple targets into account is a trend towards  artificial general intelligence.
However, how to efficiently mitigate the training conflicts among heterogeneous data collected from different domains or tasks remains under-explored.
In this study, we explore to leverage Mixture of Low-rank Adapters (MoLA) to mitigate conflicts in heterogeneous data training, which requires to jointly train the multiple low-rank adapters and their shared backbone.
Specifically, we introduce two variants of MoLA, namely, MoLA-Grad and MoLA-Router, to respectively handle the target-aware and target-agnostic scenarios during inference.
The former uses task identifiers to assign personalized low-rank adapters to each task, disentangling task-specific knowledge towards their adapters, thereby mitigating heterogeneity conflicts.
The latter uses a novel Task-wise Decorrelation (TwD) loss to intervene the router to learn oriented weight combinations of adapters to homogeneous tasks, achieving similar effects.
We conduct comprehensive experiments to verify the superiority of MoLA over previous state-of-the-art methods and present in-depth analysis on its working mechanism. Source code is available at: \url{https://github.com/MediaBrain-SJTU/MoLA}

\end{abstract}

\section{Introduction}

\begin{figure*}[t]
    \centering
   \includegraphics[width=0.99\linewidth]{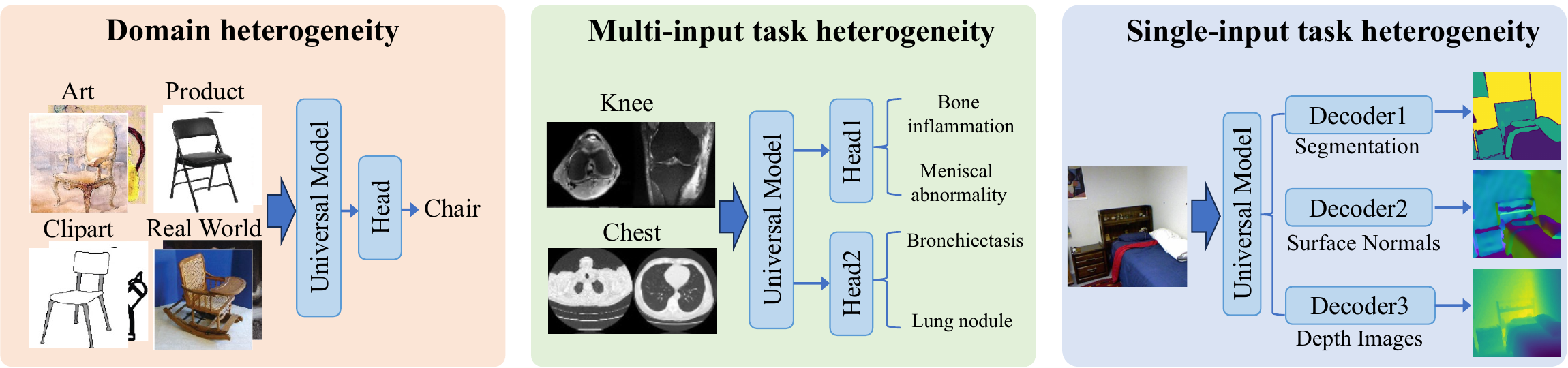}
    \caption{Three common types of heterogeneous data. {Left: Domain heterogeneity.} Each target can correspond to multiple domain heterogeneous input data, such as multi-domain training; {Middle: 
 Multi-input task heterogeneity.} Each target has its own input data, such as medical diagnosis; {Right: Single-input task heterogeneity.} Each task can have the same input data, such as scene understanding.}
    \label{fig:scene}
\end{figure*}

Diverse training data collected from different domains or tasks is often utilized to train a unified model for pursuing universal capability~\cite{sellergren2022simplified, touvron2023llama}.
However, due to the presence of heterogeneity, such unification may suffer from strong conflicts during training~\cite{aoki2022heterogeneous}, resulting in the suppression of the scale advantage of the pre-training dataset and severely impacting the performance of the model~\cite{yuan2022decentralized}. In order to address this issue, it becomes crucial to explore conflict harmony solutions for heterogeneous data training.

One close area to handle heterogeneous data is multi-task learning (MTL), which learns multiple tasks simultaneously in a single model~\cite{caruana1993multitask, crawshaw2020multi, ruder2017overview,HarmoDT,fan2024federated}. 
Standard MTL methods can be roughly categorized into two types: hard parameter sharing (HPS)~\citep{long2017learning, lu2017fully} and soft parameter sharing (SPS)~\citep{misra2016cross, liu2019end}. 
The former has not constructed task-specific branches, which results in personalized features easily diminished in a fully shared network. The contradiction of heterogeneity in the data and the complete sharing manner in the parameters limits the potential of such methods~\cite{guo2020learning, tang2020progressive}. 
The latter introduces additional parameters to enable heterogeneous feature learning and thus achieves higher performance. However, the current design lacks effective ways to control the model size, making it hard to extend to more general collaborative training scenarios~\cite{liu2019end,yao2023edge, fan2022fedskip, fan2024locally, zhang2024domain,feng2021ms}.

To pace forward the above study, we explore to combat the heterogeneity by Mixture of Low-rank Adapters (MoLA), which consists of multiple low-rank adapters~\cite{hu2021lora} attached to a shared backbone. 
Note that, previous explorations about low-rank adaptation (LoRA) mainly limit to the parameter-efficient finetuning on a pre-training model, which differs from our end-to-end aspect, as we target to jointly train the MoLA with their shared backbone. Besides, compared to the Mixture of Experts (MoE) structure~\cite{ma2018modeling,tang2020progressive}, MoLA is easier to scale up friendly due to its low-rank design. 
Generally, the intuition that we consider MoLA is two-fold: 1) the low-rank property of MoLA ensures that the increase of parameters is controllable; 2) the (primary-secondary) rank discrepancy between backbone and adapters encourages model to disentangle the shared knowledge and complementary knowledge. 
The above advantages make MoLA more flexible and capable in dealing with the heterogeneous data training.

Inspired by above analysis, we propose its two variants, MoLA-Grad and MoLA-Router, to respectively handle the target-aware and target-agnostic scenarios during inference. 
The former utilizes task identifiers to assign personalized low-rank adapters to each task, explicitly isolating parameters of different tasks. In this way, task-specific knowledge is disentangled towards their respective adapters, explicitly alleviating the heterogeneity conflicts.
The latter trains a router to dynamically manipulate the mixing weights of MoLA, achieving implicit gradient separation by assigning different combinations of the weights to different tasks. To guarantee the desired manipulation, we design a Task-wise Decorrelation (TwD) loss to force the mixing weights of different task data more heterogeneous and those of the same task data more homogeneous.
To comprehensively verify our MoLA variants, we extend ordinary MTL setting into more generalized heterogeneous data training scenarios and summarize into three types: \textit{domain heterogeneity}~\cite{torralba2011unbiased, venkateswara2017deep}, \textit{multi-input task heterogeneity}~\cite{mei2022radimagenet,yang2021medmnist} and \textit{single-input task heterogeneity}~\cite{silberman2012indoor,cordts2016cityscapes} (as shown in Fig.~\ref{fig:scene}). 
Then, we conduct extensive experiments in these three scenarios to verify the general effectiveness of MoLA.
In a nutshell, our contributions can be summarized as the following:
\vspace{-0.1cm}
\begin{itemize}[itemsep=1pt,parsep=1pt,topsep=1pt,partopsep=1pt]
\item We propose to utilize MoLA to mitigate conflicts in heterogeneous data training. 
By introducing task-specific low-rank parameters, MoLA achieves parameter isolation between different tasks, thereby separating heterogeneous gradients to avoid conflicts between tasks.
\item We propose two variants of MoLA, namely, MoLA-Grad and MoLA-Router, which use task identifiers and the router intervened by our TwD loss respectively, to effectively construct different train space for different tasks, explicitly or implicitly mitigating the conflicts.
\item We present in-depth analysis on the training of MoLA from the perspectives of principal component changes and eigenvalue distributions, and conduct extensive experiments on three common heterogeneous scenarios to fully validate the general superiority of MoLA.
\end{itemize}
\vspace{-0.2cm}


\begin{figure*}[t]
    \centering
    \includegraphics[width=0.98\linewidth]{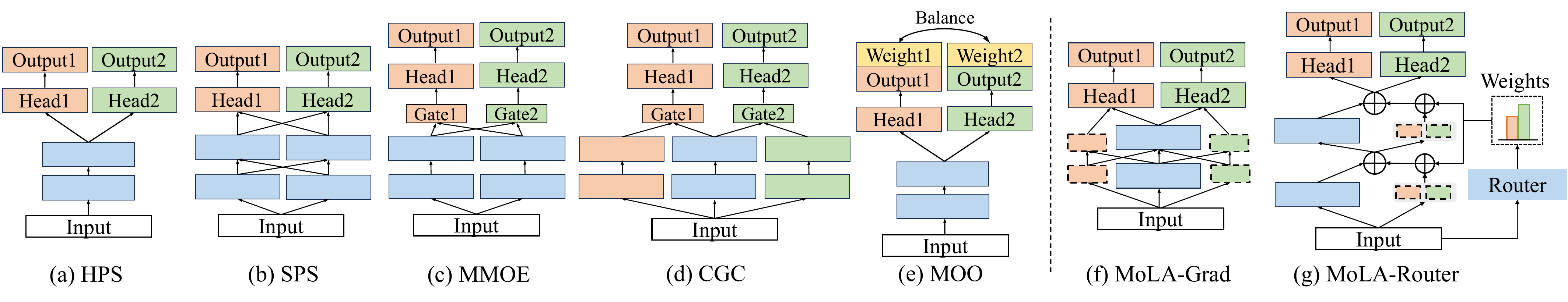}
    \vspace{-0.3cm}
    \caption{Blue rectangles represent shared modules and orange and green rectangles represent task-specific modules. The small dashed rectangles represent low-rank adapters. $\oplus$ calculates the weighted sum of MoLA  based on the output of the shared router.}
    \vspace{-0.2cm}
    \label{fig:method}
\end{figure*}

\section{Related work}
\label{related_work}
\subsection{Multi-Task Learning}

MTL methods can be roughly categorized into two types, namely hard parameter sharing (HPS)~\citep{caruana1993multitask} and soft parameter sharing (SPS)~\citep{duong2015low}. HPS methods~\citep{long2017learning, lu2017fully} share feature extractors across all tasks and can use multi-objective optimization (MOO)~\citep{kendall2018multi, chen2018gradnorm, sener2018multi} to further adjust the weights of different losses. These methods have scale invariance to a large number of tasks but easily bias towards the tasks with strong signals.
SPS methods~\citep{misra2016cross, liu2019end, wallingford2022task} utilizes the information interaction among multiple feature extractors to enhance the performance of each task, but the parameter size significantly increases with the number of tasks.
Recently,  Mixture of Experts (MoE) based MTL~\citep{shazeer2017outrageously, fan2022m3vit, aoki2022heterogeneous} has introduced the gate function to assign different parameter combinations to different tasks, demonstrating impressive performance. However, it faces the same problems as SPS, namely a large number of parameters and difficulty in scaling up training. In addition, the role played by the experts in MoE is not easily interpreted.

\subsection{Low-Rank Adaptation}

Low-Rank Adaptation (LoRA)~\cite{hu2021lora} is a parameter-efficient fine-tuning method, which has been theoretically proved able to express large models with limited LoRA-ranks~\citep{zeng2023expressive}. 
Benefiting from such capacity, LoRA has emerged as a prevalent technique for adapting foundation models to specific downstream tasks~\citep{zhang2023lora, luo2023lcm, li2023loftq, zhou2024low}. 
Previous researches preliminarily discover the potential application of LoRA on multi-task learning. \cite{chavan2023one} employs a generalized prompt module to optimize pre-trained model weights and adjust intermediate activations for better flexibility. 
\citet{huang2023lorahub} investigates LoRA composability for cross-task generalization to achieve adaptable performance on unseen tasks. 
\cite{gou2023mixture} fine-tunes multiple cluster-conditional low-rank experts to ensure zero-shot generalization across downstream tasks. \citep{dou2023loramoe} introduces LoRAMoE as a plugin version of MoE,  but its backbone and LoRA still work separately.
In this paper, we would like to make a step further to adapt LoRA to various heterogeneous data, and we are among the first attempts to include LoRA during the training procedure.

\section{Proposed Method}

\subsection{Preliminary}
Assuming our heterogeneous dataset consists of $T$ tasks, these $T$ tasks can correspond to any one  of the heterogeneous types shown in Fig.~\ref{fig:scene}. 
We define the weight of any convolution layer in the backbone as $\mathbf{W_0}\in \mathbb{R}^{C^{\text{out}}\times C^{\text{in}}\times k\times k}$, where $C^{\text{out}}, C^{\text{in}}, k$ represent the number of output channels, the number of input channels, and the kernel size respectively. 
Let $h\in \mathbb{R}^{b\times C^{\text{in}}\times H\times W}$ and $g\in \mathbb{R}^{b\times C^{\text{out}}\times H \times W}$ represent the input and output features respectively, where $b$ is the batch size, $H$ and $W$ represent the height and width of the feature maps. Then, the original convolution operation is $g=\mathbf{W_0}h$. $h_t$ and $g_t$  represent the input and output features belonging to the $t$-th task in current mini-batch.
$\mathbf{M} \in  \mathbb{R}^{b\times T}$ is the task identifier matrix where the $i$-th  row corresponds the one-hot task identifier of the $i$-th sample.

\subsection{Motivation}

The field that is most closely related to our goal is MTL. Thus, we select and compare different types of representative methods in MTL (as shown in Fig.\ref{fig:method}) to illustrate our motivation for proposing MoLA. 
HPS is the most straightforward MTL method. Its advantage lies in the fact that the number of parameters does not significantly increase with the increase of tasks. However, its drawback is that it has too few parameters available for learning task-specific features, which may result in the inability to fit the complete distribution of heterogeneous dataset. 
The same issue also applies to MMO. With the increase in the number of tasks, the parameter size of SPS significantly increases, making it challenging to apply in large-scale heterogeneous training. 

The MoE and gating functions provide an alternative way to alleviate heterogeneity conflicts. 
MMoE learns specific gating functions for each task to conduct different heterogeneous features from multiple shared extractors. CGC further enhances the ability to extract heterogeneous features by dividing the feature extractors into shared and task-specific components. However, both approaches still suffer from high scaling-up cost and the knowledge disentanglement to mediate conflicts cannot be well guaranteed as expected.

To this end, we introduce Mixture of Low-rank Adapters (MoLA), which connects multiple low-rank adapters in parallel to the weights of the backbone. 
By using task identifiers or trainable routers, MoLA combines different training parameters for different tasks. The isolation of parameters separates the gradients of different tasks, effectively mitigating conflicts in heterogeneous training.
The low-rank adapters significantly reduce the number of introduced parameters, enabling large-scale heterogeneous training. Furthermore, compared to the experts in MoE, our low-rank adapters, constrained by low-rank structures, are better at capturing task-specific features (see Section~\ref{discussion}). The fusion with the parameters of the backbone network helps maintain that the resulted structure has the sufficient representation capacity.

\subsection{Mixture of Low-Rank adapters}

For the convolution layer that requires to configure MoLA,  we assign two low rank factors $\mathbf{B_i}\in \mathbb{R}^{C^{\text{out}}k\times rk}$ and $\mathbf{A_i} \in \mathbb{R}^{rk\times C^{\text{in}}k}$ for each adapters, where $r$ represents the rank and $i\in \{1,...,E\}$, $E$ is the number of adapters defined by users.
Then, the convolution operation can be transformed into 
\begin{equation}
\setlength{\abovedisplayskip}{1pt}
\setlength{\belowdisplayskip}{1pt}
    g_t = (\mathbf{W_0}+\sum^{E}_{i=1}\alpha_i\mathbf{B_i}\mathbf{A_i}) h_t, 
    \label{eq1}
\end{equation}
where, for brevity, we omit the reshape operation, and $\alpha_i$ denotes the contribution weight of the $i$-th low rank adapter. 

Eqn~\ref{eq1} is the basic form of MoLA which can use different weighted combination of $E$ low-rank adapters and a shared $W_0$ to allocate different parameter sets for different tasks. 
By isolating parameters to separate the gradients of different tasks, heterogeneous conflicts can be alleviated. Intuitively, gradient separation can be divided into explicit and implicit types, corresponding to target-aware and target-agnostic scenarios respectively, \textit{i.e.}, MoLA-Grad and MoLA-Router.

\subsection{MoLA-Grad}
For the target-aware scenario, MoLA-Grad can utilize task identifiers to explicitly  separate gradients of heterogeneous data, alleviating the training conflicts. 
MoLA-Grad assigns the specific low-rank adapter to each task and requires the adapter to compute only the gradients for the corresponding task, thereby allowing the gradients to be explicitly separated.
In this case, we set $E=T$, $\alpha_t=1$ and  $\alpha_{i\ne t}=0$.

\textbf{Consumption Reduction.} Since the shared $W_0$ involve in the computation of all tasks, in order to avoid redundant computation and excessive memory usage, we adopt the idea of group convolution to optimize the calculation process of convolution operations.
Specifically, for each convolution with MoLA-Grad,  the output feature $g$  can be computed by
\begin{equation}
\begin{aligned}
    g & =(\mathbf{W_0}+\mathbf{B_1}\mathbf{A_1})h_1\cup\cdots\cup (\mathbf{W_0}+\mathbf{B_T}\mathbf{A_T})h_T \\
    &=(\mathbf{W_0}+  \mathbf{BA} \circ \mathbf{M}) h =\mathbf{W^{\prime}} h,
\end{aligned}
\label{group_conv}
\end{equation}
where $\mathbf{{BA}} = \mathbf{B_1}\mathbf{A_1} \cup ...\cup \mathbf{B_T}\mathbf{A_T}$, $\cup$ means the concatenation operation and $\circ$ means the element-wise multiplications through broadcasting.
Then we reshape the aggregated weight $\mathbf{W^{\prime}} \in \mathbb{R}^{b\times C^{\text{out}}\times C^{\text{in}} \times k \times k}$  to $\mathbb{R}^{bC^{\text{out}}\times C^{\text{in}}\times k\times k}$,  $h$ to $\mathbb{R}^{1\times bC^{\text{in}}\times H\times W}$,  and set the group number to $b$. In this way, Eqn.~\eqref{group_conv} will be a standard group convolution, which can be easily implemented in existing deep learning libraries.

\subsection{MoLA-Router}
\label{router}
For the target-agnostic scenario, MoLA-Router train a router for dynamically generating task-specific weighted combination to implicitly separate heterogeneous gradients. 
In this case,  we only require $\sum_{i=1}^E\alpha_i=1$.
Although MoLA-Router cannot achieve explicit gradient separation like MoLA-Grad, it has a wider range of applicable scenarios and promotes collaboration among low-rank adapters.

The router can be implemented by any structure and applied at any layer. 
Here, for simplicity, we stack three residual blocks $\phi(\theta)$ and a gate function $g(\eta )$ to build a shared router for all layers with MoLA. Then, for input samples $x$,  the contribution weights output by the router can be written as 
\begin{equation}
       \vec{\alpha} = \mathrm{softmax}(g(\phi(\theta,x),\eta)),\\
\end{equation}
where $\vec{\alpha} \in \mathbb{R}^{b\times E}$. However, unconstrained routers may not be able to output desired mixing weights for different tasks, resulting in the entanglement of parameters from different tasks and thus failing to alleviate heterogeneous conflicts. 
To prevent insufficient discriminability, we first introduce a trainable MLP $\varphi(\cdot)$ to map $\vec{\alpha}_i$ to a higher dimension space, \textit{i.e.}, $\omega=\varphi(\vec{\alpha})$. Then, we propose a Task-wise Decorrelation (TwD) loss $\mathcal{L}_\mathrm{TwD}$ to supervise the learning of mixing weights, which aims to promote similarity in contribution weights among data from the same task while emphasizing distinctiveness in contribution weights between data from different tasks. 
Specifically, our $\mathcal{L}_\mathrm{TwD}$ can be written as 
\begin{equation}
\setlength{\abovedisplayskip}{1pt}
\setlength{\belowdisplayskip}{1pt}
\mathcal{L}_\mathrm{TwD}=-\sum_{i=1}^b\sum_{j=1}^b\mathbbm{1}_{i\neq j}\cdot\mathbbm{1}_{t_i=t_j}\log\frac{\mathrm{e}^{{{\omega}}_i^\top{{\omega}}_j/\tau}}{\sum_{k=1}^b\mathbbm{1}_{i\neq k}\mathrm{e}^{{{\omega}}_i^\top{{\omega}}_k/\tau}},
\nonumber
\end{equation}
where $t_i, t_j$ represent the task identifiers (unavailable during inference) of the $i$-th and the $j$-th input sample, and $\mathbbm{1}$ is the indicator function, $\tau$ is temperature with the default value 1.

\subsection{Loss Function}
The loss function of using MoLA during heterogeneous data training is the same as that of multi-task training. Let $\mathcal{L}_i$ denote the loss function of the $i$-th task, and the total loss is
\begin{equation}
\setlength{\abovedisplayskip}{1pt}
\setlength{\belowdisplayskip}{1pt}
    \mathcal{L}_{total} = \sum_{i=1}^T\frac{1}{T}\mathcal{L}_i+\beta\mathcal{L}_\mathrm{TwD},
\end{equation}
where $\beta$ is a hyper-parameter, and is  nonzero when using the MoLA-Router paradigm for target-agnostic scenarios.

\section{Discussion}
\label{discussion}
\begin{figure}[t!]
    \centering
   \includegraphics[width=0.99\linewidth]{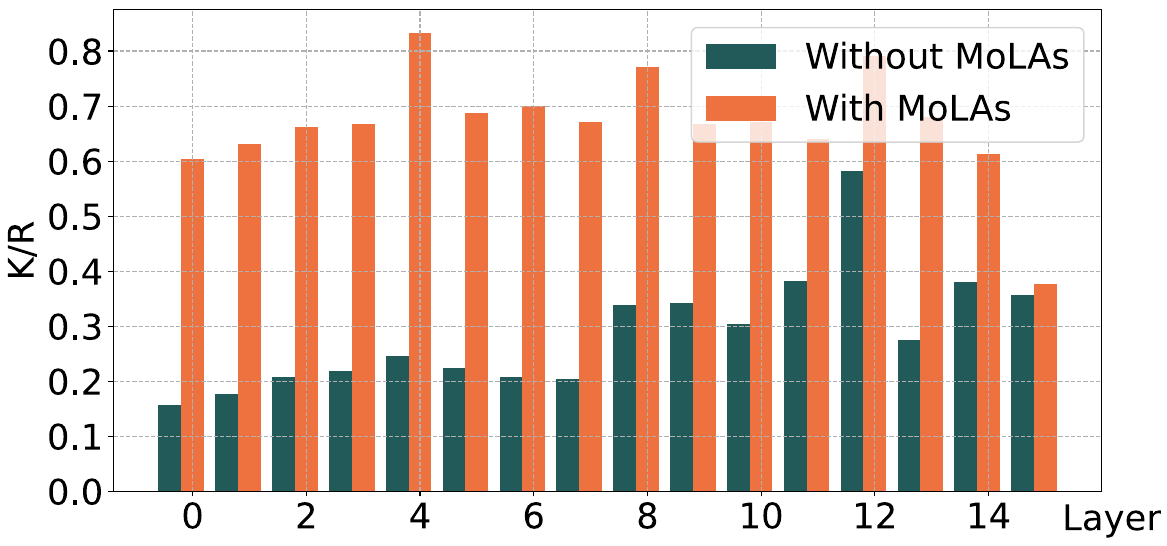}
   \vspace{-0.3cm}
    \caption{The proportion of principal component eigenvectors in the model's weight matrix. After using MoLA, the proportion significantly increases, indicating that more eigenvectors are utilized, which is beneficial for expressing task-specific directions.}
    \label{fig:bar}
    \vspace{-0.5cm}
\end{figure}
In this section, we will analyze the impact of MoLA on model training from the perspectives of principal components and eigenvalue distributions, and summarize the differences between MoLA and the original LoRA at last.

\begin{figure*}[t!]
    \centering
   \includegraphics[width=0.93\linewidth]{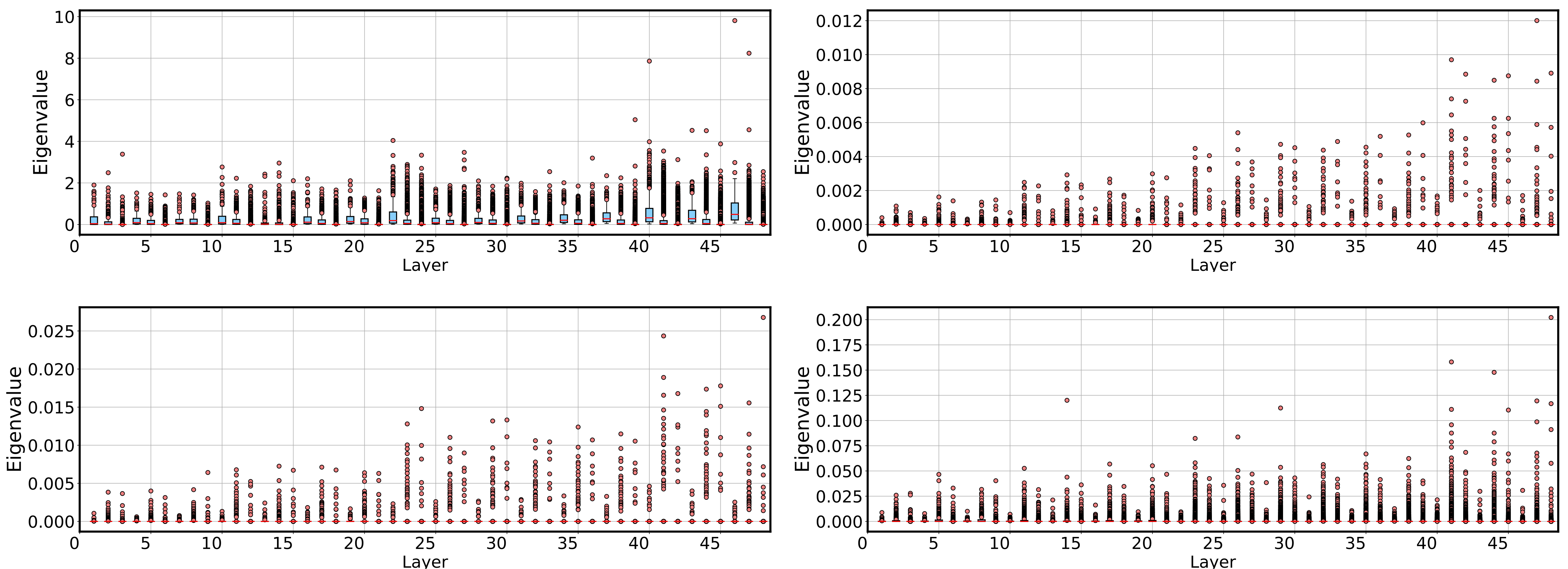}
   \vspace{-0.25cm}
    \caption{The box-plot of eigenvalue distribution of weights at different layers. The outlier points above boxes correspond to the relatively large eigenvalues, indicating that its corresponding eigenvector plays an important role in feature extraction.
    The top-left corresponds to the parameters $W_0$ in the backbone.  The top-right and bottom-left correspond to the parameter combinations of different low-rank adapters with  $W_0$. The bottom-right corresponds to the parameter combination of a low-rank adapter with a higher rank and $W_0$.
    }
    \label{fig:box}
    \vspace{-0.1cm}
\end{figure*}

\par{\noindent \bf Changes in principal components.} 
We define the principal components of the weight as the top $K$ largest singular values and their corresponding eigenvectors. Given a principal component factor $\alpha$ ($\alpha$=0.99 in our analysis), and the rank $R$ of the weight, $K$ can be obtained by
$\sum_{i=1}^{K}\sigma_i\ge\alpha\sum_{i=1}^{R}\sigma_i$, where $K$ is the minimum value that satisfies the inequality, and $\{\sigma_1,...\sigma_R\}$ represents the sorted singular values. We compare the proportion of principal component eigenvectors $K/R$ using and not using MoLA in Fig.~\ref{fig:bar}.

It can be observed that when not using MoLA, for the majority of layers, the value of $K/R$ is less than 0.4. This means that in feature extraction, less than half of the eigenvectors contribute to over 99\% of the effect. Furthermore, since all the parameters of the feature extractor are shared across all heterogeneous tasks, the focus is more on extracting shared information from heterogeneous data, which may only require the participation of partial eigenvectors.

After using MoLA, there is a significant increase in 
$K/R$, indicating that more eigenvectors play a primary role in feature extraction. In other words, the introduction of MoLA allows for the extraction of more task-specific heterogeneous features, thus requiring the involvement of a greater number of eigenvectors for representation. 

In conclusion, one of the main impacts brought by MoLA for training can be summarized as providing conditions for extracting heterogeneous features by introducing task-specific parameter combinations.

\par{\noindent \bf Changes in the eigenvalue distribution.} 
We compare the eigenvalue distributions of the  weights in Fig.~\ref{fig:box} and our findings can be summarized as follows: 
1) By comparing the top-left with the others, it can be observed that MoLA can significantly reduce the maximum outlier value of the eigenvalues. This indicates that MoLA can effectively weaken the dominant position of certain eigenvectors, thereby encouraging more eigenvectors to participate in capturing heterogeneous features.
2) By comparing the top-right and bottom-left, it can be observed that the combination of $W_0$ with different low-rank adapters leads to different distributions of eigenvalues. This indicates that different low-rank adapters in MoLA have different tendencies in feature extraction, corresponding to capturing heterogeneous features for different tasks.
3) By comparing the top-right/bottom-left with the bottom-right, it can be observed that as the rank increases, the range of eigenvalue distribution also increases, but it is still much smaller than the top-left. This indicates that appropriately increasing the rank can avoid the dominance of certain eigenvalues and enhance the expressive power of some eigenvalues.

\par{\noindent \bf The difference from the original LoRA.} 
The differences can be summarized as follows: 
1) \emph{different learning stages.} LoRA is typically used for adapting pre-trained  models to downstream tasks, while our MoLA is used to train from scratch together with the backbone;
2) \emph{different impacts on training.} The analysis in \cite{hu2021lora} indicates that  LoRA can significantly amplify a small number of eigenvalues, thereby emphasizing task-relevant eigenvectors. Instead, our MoLA significantly reduce the maximum eigenvalues to capture more heterogeneous information, alleviating training conflicts;
3) \emph{different usage scenarios.} LoRA is commonly applied to specific downstream tasks, while our MoLA also should consider learning of the mixing weights in the downstream target-agnostic scenario.

\definecolor{lightgreen}{rgb}{0.68, 0.93, 0.68}

\begin{table*}[th]\footnotesize
\caption{Comparisons on RadImageNet (Multi-input task heterogeneity), we report the performance of each task and their average.}
\centering
\resizebox{0.95\textwidth}{!}{
\setlength{\tabcolsep}{1mm}{
\begin{tabular}{l|ccccccccccc|c}
\toprule[1.5pt]
  & Lung $\uparrow$ & Abdomen $\uparrow$ & Thyroid $\uparrow$ & Abdomen $\uparrow$ & Knee $\uparrow$ & Shoulder $\uparrow$ & Spine $\uparrow$ & Ankle $\uparrow$ & Abdomen $\uparrow$  & Brain $\uparrow$ & Hip $\uparrow$  & Avg $\uparrow$ \\
\midrule
{Single-Task }  & 76.42 & 33.94 & 91.55 & \textbf{69.17} & \underline{49.32} & 41.80 & 20.62 & \textbf{20.31} & 65.99 & 83.88 & 51.05  & 54.91\\
{Uniform }  & 34.34 & \underline{41.33} & 69.85 & 22.87 & 34.71 & 27.86 & 20.26 & 12.88 & 60.01 & 75.16 & 24.13  & 38.49\\
\midrule
{HSP}  & 77.16 & 37.45 & 91.73 & 68.43 & 46.47 & 42.72 & 20.85 & 18.17 & 71.13 & {84.67} & \underline{55.16} & 55.81 \\
MGDA &70.61 &31.74 &\textbf{94.22} &67.46 &43.59 &45.18 &\underline{23.69} & 17.67&73.86 & 84.20&53.62 &55.08\\
PCGrad &69.34 & \textbf{43.15}& 75.60&67.77 &42.59 &39.24 &14.20 &14.38 &34.52 &69.36 &46.26 &46.95\\
CAGrad & 71.73 & 22.71 &91.83 & 62.70& 42.30&42.84 &\textbf{24.42} & 18.99& \underline{75.07}&83.83 &50.51 &53.36\\
Aligned-MTL  &62.59 &33.90 &\underline{92.96} &67.18 &44.33 &\underline{45.41} &23.41 & 18.28& 71.68& \textbf{84.93}&54.84 &54.50\\
\midrule
Cross-Stitch &80.71 &26.67 &92.73 & 66.03&44.25 &44.13 & 23.01& 17.92& 65.22& 74.37& 47.51&52.96\\
MMoE & \textbf{81.75}&38.65 &83.27 &67.27 &44.35 &43.84 &16.42 &13.11 & 47.64& 77.64& \textbf{55.63}&51.78\\
DSelect-K & 77.48& 35.34& 91.62& 67.08& 45.89& 42.22&19.50 &15.49 &73.39 &79.74 &53.85 &54.69\\
CGC & 75.47& 28.12& 86.26& 67.67& 46.02& 42.16& 15.09& 15.81& 24.93& \underline{84.88}& 53.04& 49.04\\
LTB & 68.63& 40.69& 88.99&68.09 & 45.69& \textbf{45.61}& 23.13&18.79 &\textbf{75.39} &84.39 &53.56 &55.72\\
\midrule
\rowcolor{gray!10}
MoLA-Router & 78.94&36.38 & 91.76& 68.05& 48.41& 43.03& 23.26&18.37 & 68.65& 84.56& 54.93&\underline{56.03}\\
\rowcolor{gray!10}
MoLA-Grad  & \underline{80.72} & 34.54 & 92.18 & \underline{68.87} & \textbf{50.18} & 43.41 & 22.06 & \underline{19.76}  & 69.10 & 84.67 & \textbf{55.63}  & \textbf{56.47}  \\
\bottomrule
\end{tabular}}}
\vspace{-0.6cm}
\label{rad}
\end{table*}

\begin{table}[t!]
\caption{Comparisons on VLCS (domain heterogeneity). }
    \centering
    \resizebox{0.44\textwidth}{!}{
    \setlength{\tabcolsep}{0.7mm}{
    \begin{tabular}{l|cccc|c}
    \toprule[1.5pt]
  & {Domain-V} & {Domain-L}  & {Domain-C} & {Domain-S}  &Avg $\uparrow$ \\ 
 \midrule
 Single-Task & 84.32 & 75.40 & 100.0 & 78.70  & 84.60 \\ 
 Uniform & 84.75& 72.73& 100.0& 76.09 &83.39 \\
 \midrule
 MMD-AAE& 84.32&69.52 &100.0 &80.43   &83.57 \\
 SelfReg & 81.36& 71.65&100.0 &\underline{82.17 }  & 83.80\\
 EQRM & 83.9& 67.38& 100.0& 79.57  &82.71 \\
 DANN &49.15 & 57.75&60.00 & 55.65  &55.64 \\
 \midrule
 HPS &85.59 &74.33 & 100.0&80.00 & 84.98 \\
 Cross-Stitch &\underline{86.44} &\underline{78.07} & 100.0& 77.83& 85.59\\
 MMoE &83.05 &74.33 &100.0 &79.57& 84.24\\
 DSelect-K &83.90 &75.40 & 100.0&80.87 & 85.04\\
 CGC & 83.90& 72.73& 100.0&80.43 & 84.27\\
 LTB & 80.93& 75.94& 100.0& 79.13& 84.00\\
 \midrule
 \rowcolor{gray!10}
 MoLA-Router &85.59 &\textbf{79.14} & 100.0&80.87 &\underline{86.40}\\
  \rowcolor{gray!10}
MoLA-Grad &\textbf{87.29} &74.87 & 100.0& \textbf{83.48} &\textbf{86.41}\\
    \bottomrule[1.5pt]
    \end{tabular}}}
    \label{tab:vlcs}
\vspace{-0.6cm}
\end{table}

\section{Experiments}

In this section, we conduct experiments on a total of five datasets from three different heterogeneous types to validate the effectiveness of our approach. Corresponding datasets and competing methods are introduced as follows.

\textbf{Datasets.} 
For domain heterogeneity, we use \textbf{VLCS}~\citep{torralba2011unbiased} and \textbf{Officehome}~\citep{venkateswara2017deep} datasets;
For Multi-input task heterogeneity, we use \textbf{RadImageNet}~\citep{mei2022radimagenet} and \textbf{MedMNISTV2}~\citep{yang2021medmnist} datasets;
For Single-input task heterogeneity, we use 
\textbf{NYUv2}~\citep{silberman2012indoor}.
A detailed description of the dataset is provided in the appendix.

\textbf{Competing methods.} We compare our methods with the following MTL methods and baselines: Single-task~\citep{sun2020adashare}, HSP~\citep{wei2021hps}, Cross-Stitch~\citep{misra2016cross}, MMoE~\citep{ma2018modeling}, CGC~\citep{cheng2016cgc}, LTB~\citep{guo2020learning}, DSelect-k~\citep{hazimeh2021dselect}, MMD-AAE~\citep{li2018domain}, SelfReg~\citep{kim2021selfreg}, EQRM~\citep{eastwood2022probable}, DANN~\citep{muralidhar2018incorporating}, MGDA(-UB)~\citep{sener2018multi}, IMTL~\citep{liu2021towards}, Nash-MTL~\citep{navon2022multi}, MoCo~\citep{he2020momentum}, Aligned-MTL(-UB)~\citep{senushkin2023independent}, RLW~\citep{lin2021reasonable}, GradNorm~\citep{chen2018gradnorm}, GradDrop~\citep{chen2020just}, PCGrad~\citep{yu2020gradient}, GradVac~\citep{wang2020gradient}, DWA~\citep{liu2019end}and CaGrad~\citep{liu2021conflict}.
The description and classification of above methods is provided in the appendix.

\subsection{Domain Heterogeneity}

The comparison results on domain-heterogeneity datasets are shown in Tab.~\ref{tab:vlcs} and Tab.~\ref{tab:officehome}. It can be observed that the performance of Uniform does not always surpass Single-Task, indicating that domain heterogeneity can lead to training conflicts.
Domain-alignment based methods perform significantly better than baselines and MTL methods on Officehome, but not on VLCS. This indicates that the performance of domain-alignment based methods requires sufficient data support.
In addition, MTL based methods do not demonstrate the expected advantages on both datasets. In fact, there have been few works in the MTL field that specifically address domain-heterogeneity.
Our MoLA-Grad achieves the best performance on both datasets, which demonstrates the effectiveness of the explicit gradient separation in mitigating conflicts caused by domain heterogeneity. 
However,  for the domain-heterogeneity scenario, MoLA-Grad requires task identifiers to be provided during testing in order to use the corresponding low-rank adapters. Fortunately, our MoLA-Router achieved the second highest performance, validating the effectiveness of MoLA.


\begin{table}[t!]
\caption{Comparisons on Officehome (domain heterogeneity).}
    \centering
    \resizebox{0.44\textwidth}{!}{
    \setlength{\tabcolsep}{0.7mm}{
    \begin{tabular}{l|cccc|c}
    \toprule[1.5pt]
  & {Domain-P}  & {Domain-A}  & {Domain-C}  & {Domain-R}  &Avg $\uparrow$ \\ 
 \midrule
 Single-Task & 87.78 & 70.33 & 72.73 & 81.80  & 78.16 \\
 Uniform & 90.89& 71.14& 79.25& 85.48 & 81.69\\
  \midrule
 MMD-AAE &90.22 &74.80 &\underline{80.65} &83.64   &82.33 \\
 SelfReg &91.11 &76.42 &\underline{80.65} & 84.56  & 83.19\\
 EQRM &90.44 &\textbf{77.64} &\underline{80.65} & \underline{86.17} & 83.73\\
 DANN &89.56 &73.17 &78.78 &82.36   & 80.94\\
 \midrule
 HPS &86.44 & 71.95& 74.36&79.95 & 78.18\\
 Cross-Stitch & 88.44&65.45 &74.59 &79.26 & 76.94\\
 MMoE &89.56 &66.67 &75.76 &78.80& 77.70\\
 CGC & 90.22& 69.11& 76.92& 78.80& 78.76\\
 DSelect-K & \underline{91.33}&72.76 &78.32 &81.11 & 80.88\\
 LTB & 84.22& 61.79& 71.79& 73.04& 72.71\\
 \midrule
  \rowcolor{gray!10}
 MoLA-Router &\textbf{93.33} &77.05 &\textbf{80.89} &84.10 &\underline{83.84}\\
  \rowcolor{gray!10}
  MoLA-Grad &\underline{91.33} &\underline{77.24} &\textbf{80.89} & \textbf{86.64}&\textbf{84.02}\\
    \bottomrule[1.5pt]
    \end{tabular}}}
    \vspace{-0.6cm}
    \label{tab:officehome}
\end{table}

\begin{table*}[th!]\footnotesize
\centering
\caption{Comparisons on MedMNISTV2 (Multi-input task heterogeneity), we report the performance of each task and their average.}
\resizebox{0.97\textwidth}{!}{
\setlength{\tabcolsep}{3mm}{
\begin{tabular}{l|ccccccccc|c}
\toprule[1.5pt]
   & Colon $\uparrow$ & Retinal $\uparrow$ & OrganC $\uparrow$ & Cell $\uparrow$& Breast $\uparrow$  & Skin $\uparrow$ & OrganA $\uparrow$ & OrganS $\uparrow$ & Chest $\uparrow$   & Avg $\uparrow$ \\
\midrule
{Single-Task } & 84.53 & 78.40 & 89.65 & \textbf{96.81} & \underline{85.26} & 73.97 & 92.90 & 77.43 & 85.42 & 84.93\\
{Uniform} & 87.41 & 77.40 & 23.51 & 50.37 & 84.62  & 12.92 & 18.64 & 18.90 & 86.22 & 51.11\\
\midrule
{HSP} &87.74 & 74.00&\underline{90.22} & \underline{95.03}& \textbf{87.82}&73.72 & 92.28& 76.50& 80.45& 84.19\\
MGDA & 85.11 &73.10 &71.48 &74.45 & 35.90&63.74 &80.89 & 54.21& \underline{88.78}& 69.74 \\
PCGrad & 87.56& 75.60& 89.32& 94.15& 81.41& 72.37& 92.23&76.21 &83.97 & 83.65 \\
CAGrad& 84.94& 65.30& 88.27& 92.98& \underline{85.26} & 71.62& 91.46&73.85 &81.73&81.71 \\
\midrule
Cross-Stitch &86.25 & 76.40& 89.65&94.88 &82.05 &\textbf{75.11} &92.98 & 76.11&84.94&84.26\\
MMoE &86.32& 76.50& 89.85& 94.68& 76.92& \underline{74.26}&92.12 &73.81 &85.58&83.34 \\
MTAN &58.89 &73.90 &88.93 &92.90 & 84.62& 74.01& 93.07& 75.55 &77.08&79.88  \\
CGC &\textbf{90.10} &\underline{80.40} &86.13 &93.42 &83.33 &73.27 &90.96 &73.49 &86.22&84.15  \\
DSelect-K & 89.12& 77.10& 87.92& 94.86& 83.97& 70.92& 91.08 &74.26 &87.66 &84.10\\
\midrule
 \rowcolor{gray!10}
MoLA-Router & \underline{89.22}& 75.40& \textbf{90.76}& 94.01& 83.97&72.47 & \underline{93.43}& \textbf{79.44}& \textbf{91.03}&\textbf{85.53}\\
 \rowcolor{gray!10}
MoLA-Grad  &88.08 & \textbf{81.00}& 89.85& 93.95&  \textbf{87.82}& 74.11 & \textbf{93.78}& \underline{77.79}&82.85&\underline{85.47} \\
\bottomrule[1.5pt] 
\end{tabular}}}
\vspace{-0.35cm}
\label{medmnist}
\end{table*}

\begin{table*}[th!]
\caption{Comparisons on NYUv2 (Single-input task heterogeneity).}
    \centering
    \resizebox{0.94\textwidth}{!}{
    \setlength{\tabcolsep}{4mm}{
    \begin{tabular}{l|ccccccccc|c}
    \toprule[1.5pt] 
  & \multicolumn{2}{c}{Segmentation $\uparrow$} & \multicolumn{2}{c}{Depth $\downarrow$}  & \multicolumn{5}{c}{Surface normal $\downarrow$}  &  \\ 
& && &  & \multicolumn{2}{c}{Angle Dist. $\downarrow$} & \multicolumn{3}{c}{Within $t^{\circ}$  $\uparrow$} & $\Delta$ m\% $\downarrow$\\ 
 Method & mIoU & Pix Acc & Abs. & Rel. & Mean & Median & 11.25 & 22.5 & 30 & Task-weighted \\
 \midrule
 Single-Task &49.37 &72.03 &0.52 & 0.24& 22.97&16.94 & 0.34& \underline{0.62}& \underline{0.73}&- \\
 \midrule
 HSP & 45.21& 69.70& 0.49& 0.21& 26.10& 21.08& 0.26& 0.52& 0.66&  4.72\\
 RLW& 46.19& 69.71&0.46 & 0.19& 26.09& 21.09& 0.27& 0.53& 0.66& 1.73\\
 DWA& 45.83& 69.65& 0.50& 0.22& 26.10& 21.27& 0.26& 0.52& 0.66&  5.61\\
 MGDA& 40.96& 65.80& 0.54& 0.22& 23.36& 17.45& 0.33& 0.61& 0.72&  4.24\\
 MGDA-UB&41.15 &65.10 &0.53 &0.22 &23.42 &17.60 & 0.32& 0.60& 0.72&  4.40\\
 GradNorm&45.63 & 69.64& 0.48& 0.20& 25.46& 20.06& 0.28& 0.55& 0.67& 2.18\\
 GradDrop& 45.69& 70.13& 0.49& 0.20&26.16 & 21.21& 0.26&0.52 & 0.65&  3.92\\
 PCGrad& 46.37& 69.69& 0.48& 0.20& 26.00& 21.05& 0.26& 0.53& 0.66&  3.17\\
 GradVac& 46.65& 69.97& 0.49& 0.21& 25.95& 20.88& 0.27& 0.53& 0.66& 3.75\\
 CAGrad& 45.46& 69.35& 0.47& 0.20& 24.28& 18.73& 0.30&0.58 &0.70  &0.13 \\
 IMTL&44.02 &68.56 &0.47 &0.19 &23.69 & 18.03& 0.32& 0.59& 0.72&  -1.02\\
 Nash-MTL&47.25 &70.38 & 0.46& 0.20& 23.95& 18.83& 0.31& 0.59& 0.71& -1.48\\
 Aligned-MTL&46.70 & 69.97& 0.46& 0.19& 24.19& 18.77& 0.30& 0.58& 0.71 &-1.55 \\
 Aligned-MTL-UB&46.47 & 69.92& 0.48& 0.20& 24.37& 18.88& 0.30& 0.58& 0.70  &0.07 \\
\midrule
 Cross-Stitch&50.06 &72.73 & 0.45&0.18 &24.52 & 18.86 &0.30 &0.57 &0.70 &  -4.42 \\
 MMoE& 51.05& 73.57& 0.49& 0.20&\textbf{22.61} &\textbf{16.33}& \underline{0.35}& \textbf{0.63}& \textbf{0.74}& -5.69 \\
 MTAN&\underline{52.07} &73.60 &0.43 &\underline{0.17} & 24.34&18.92 &\textbf{0.39} &0.57 &0.70 & -6.56 \\
 CGC&50.05 &72.92 &0.46 &0.19 & \underline{22.67}& \underline{16.72}& 0.34& \underline{0.62}& \textbf{0.74}& -6.55 \\
LTB& 46.46&70.49 &0.48 &0.19 & 24.74&18.73 &0.31 & 0.57&0.69 & -0.42\\
DSelect-K& 48.57& 70.49& 0.90& 0.37& 27.12&22.71 &0.25 &0.49 & 0.63& 29.64\\
  \midrule
   \rowcolor{gray!10}
 MoLA-Router &51.30 & \underline{73.74}& \underline{0.40}& \textbf{0.16} &24.11 &18.73 &0.31 & 0.58& 0.70& \underline{-8.18}\\
  \rowcolor{gray!10}
 MoLA-Grad & \textbf{52.79} & \textbf{75.08} &\textbf{0.39} & \textbf{0.16} & 24.00 & 18.73 &0.31 &0.58 &0.70 & \textbf{-9.17}\\
    \bottomrule[1.5pt] 
    \end{tabular}}}
    \vspace{-0.5cm}
    \label{tab:nyud}
\end{table*}

\subsection{Multi-input Task Heterogeneity}

The comparison results on multi-input task heterogeneity datasets are shown in Tab.~\ref{rad} and Tab.~\ref{medmnist}.  It can be observed that Uniform exhibits a significant performance decline compared to Single-Task, indicating that the heterogeneous conflict caused by multi-input task heterogeneity is severe. All MTL methods have significantly improved performance compared to Uniform, demonstrating the effectiveness of these methods in mitigating heterogeneous conflicts. However, the performance of these methods is difficult to surpass that of Single-Task, failing to reflect the advantages brought by large-scale data. Our MoLA-Grad and MoLA-Router achieve the highest and second-highest performance on both datasets, and their performance surpasses Single-Task. This demonstrates our MoLA can effectively mitigate this heterogeneity conflict and benefit the model from large-scale data. 
It is worth noting that in this scenario, the task identifiers required by MoLA-Grad are generally provided naturally due to the significant differences between different tasks.

\subsection{Single-input Task Heterogeneity}
The comparison results on single-input task heterogeneity datasets are shown in Tab.~\ref{tab:nyud}.
Besides task specific metrics, we follow~\cite{senushkin2023independent} and report a model performance drop relative to a single task baseline averaged over tasks: $\Delta m_{task} = \frac{1}{T}\sum_{t=1}^T\sum_{k=1}^{n_t}(-1)^{\sigma_{tk}}(M_{m, tk}-M_{b, tk})/M_{b,tk}$, where $M_{m,tk}$ denotes the performance of a model $m$ on a task $t$, measured with a metric $k$. Similarly, $M_{b, tk}$ is a performance of a single-task $t$ baseline; $n_t$ denotes number of metrics per task $t$. $\sigma_{tk}=1$ if higher values of metric is better, and $\sigma_{tk}=0$ otherwise.

It can be observed that MTL methods based on SPS and MoE outperform those based on HPS and MOO in terms of performance. This indicates that the latter may have bottlenecks in the model structure, resulting in limited benefits obtained from the relevant tasks.
Our MoLA-Grad and MoLA-Router achieved the highest and second-highest performance, respectively, further demonstrating the effectiveness of MoLA. Due to the distinct characteristics of the target tasks, in such a heterogeneous scenario, the task identifiers of MoLA-Grad are naturally provided.

\subsection{Brief Summary}
In general, for these three different heterogeneous scenarios, we found that the performance of MoLA-Grad is better than MoLA-Router, and both are superior to other comparative methods. This indicates that separating gradients through parameter isolation is a viable solution to alleviate heterogeneous conflicts. Explicit gradient separation achieves the best results but requires task identifiers, while implicit gradient separation yields slightly inferior results without being constrained by scenarios. However, in fact, due to the significant differences between heterogeneous data, task identifiers can be inferred or naturally provided, making MoLA-Grad applicable in a wide range of scenarios.

\begin{figure}[t]
    \centering
     \includegraphics[width=0.98\linewidth]{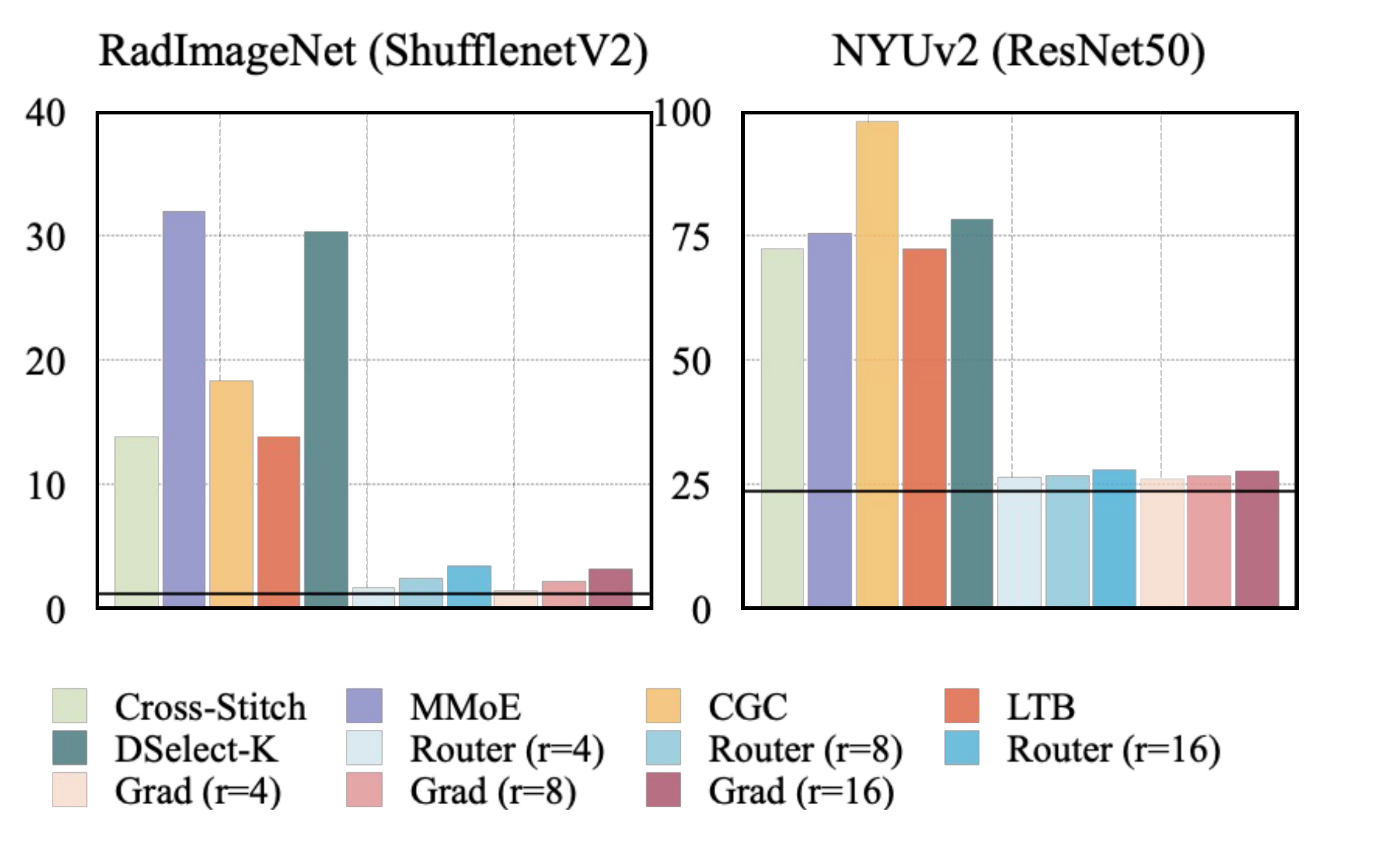}
     \vspace{-0.3cm}
    \caption{The comparison of parameter number of different methods, and the influence of rank r selection on parameter number. The vertical axis corresponds to the ``Params (M)''. The black horizontal line represents the parameter count of the single task model.}
    \vspace{-0.6cm}
    \label{fig:params}
\end{figure}

\begin{table}[t]
\caption{The impact of the rank $r$ on the performance of MoLA-router and MoLA-grad. The experiment is based on VLCS dataset.}
    \centering
    \resizebox{0.48\textwidth}{!}{
    \setlength{\tabcolsep}{1mm}{
    \begin{tabular}{l|cccc|c}
    \toprule[1.5pt]
 & {Domain-V}  & {Domain-L}   & {Domain-C}  & {Domain-S}  &Avg $\uparrow$ \\ 
 \midrule
 Single-Task & 84.32 & 75.40 & 100.0 & 78.70  & 84.60 \\ 
 Uniform & 84.75& 72.73& 100.0& 76.09 &83.39 \\
  \midrule
Router(r=1) & 83.47  & 74.33 & 100.0 & 77.39  & 83.80 \\
\rowcolor{gray!30}
Router(r=4)  &85.59 &79.14 & 100.0&80.87 &86.40\\
Router(r=8)  & 84.32  & 72.73 & 100.0 & 76.96  & 83.50 \\
Router(r=16)  & 83.90  & 75.94 & 99.00 & 76.52  & 83.84 \\
\rowcolor{gray!15}
Router(r=64)  & 83.90  & 77.54 & 100.0 & 79.57  & 85.25 \\
 \midrule
Grad(r=1) & 82.20  & 79.14 & 100.0 & 75.65  & 84.25 \\
\rowcolor{gray!30}
Grad(r=4)  &87.29 &74.87 & 100.0& 83.48 &86.41\\
Grad(r=8)  & 83.05  & 74.33 & 100.0 & 80.00  & 84.35 \\
Grad(r=16)  & 84.75  & 74.33 & 100.0 & 77.83  & 84.23 \\
\rowcolor{gray!15}
Grad(r=64)  & 83.47  & 75.94 & 100.0 & 78.70  & 84.53 \\
    \bottomrule[1.5pt]
    \end{tabular}}}
    \vspace{-0.7cm}
    \label{tab:rank}
\end{table}

\subsection{Ablation study}
\par{\noindent \bf Comparison of parameter number.}
Since our method and MTL methods based on SPS/MoE  all introduce additional training parameters, we compare their parameter number to demonstrate the parameter efficiency of MoLA. The results are shown in Fig.~\ref{fig:params}.
It can be seen that our method has a significant advantage in controlling the number of parameters, and the parameter count does not increase significantly with the increase of $r$. Compared to other methods, the parameter efficiency of our proposed MoLA provides greater potential for large-scale heterogeneous data training.

\begin{table}[t]
    \centering
    \caption{The impact of embedding MoLA at different positions in the model on performance. The experiment is based on NYUv2. We report $\Delta m_{task}\%$, and the lower value is better.}
    \resizebox{0.48\textwidth}{!}{
    \setlength{\tabcolsep}{3.5mm}{
    \begin{tabular}{c|c|c|c|c}
        \toprule[1.5pt]
         Without MoLA &  {Block1} & {Block2} & {Block3} & {Block4}  \\
        \midrule
          4.72 & 0.02 & -1.71 &  -0.07 & -9.17\\
          \midrule
          \midrule
         {Block12} &  {Block34} & {Block123} & {Block234}  & {Block1234}  \\
        \midrule
          -3.98  & -3.95 &  -0.89 & -3.26&-1.30\\
        \bottomrule[1.5pt]
    \end{tabular}}}
    \vspace{-0.25cm}
    \label{tab:position}
\end{table}

\begin{figure}[t]
    \centering
   \includegraphics[width=0.98\linewidth]{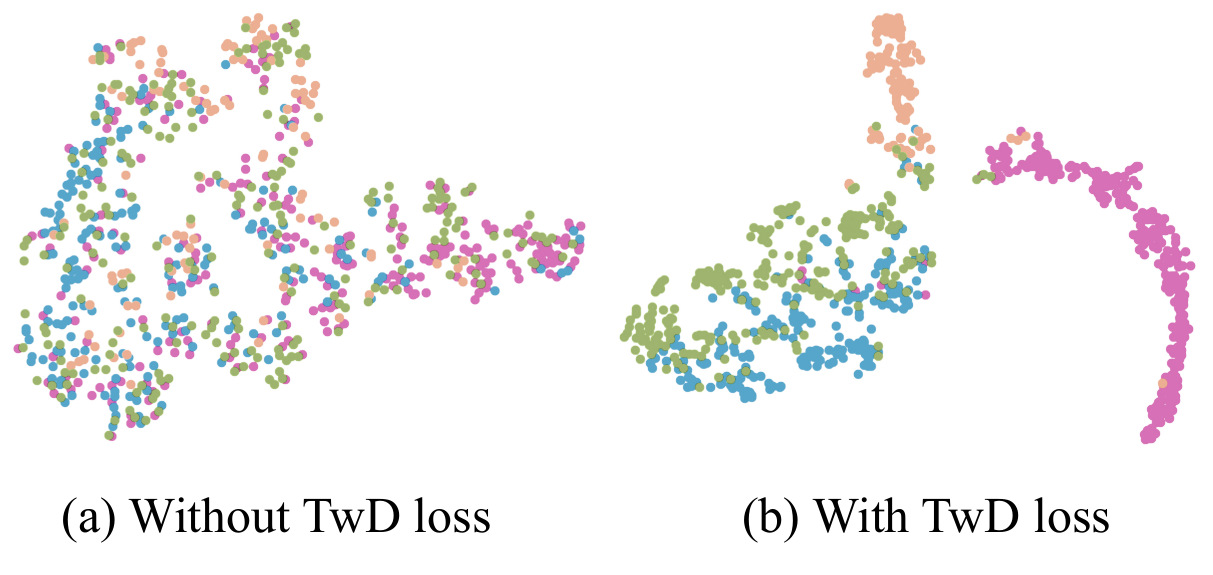}
   \vspace{-0.3cm}
    \caption{A TSNE visual comparison of the contribution weights output by the trained router with and without TwD.}
    \vspace{-0.6cm}
    \label{fig:tsne}
\end{figure}

\begin{table}[t]
\caption{Performance comparison between the ResNet structure and the Transformer structure 
 based on Officehome and VLCS.}
    \centering
    \resizebox{0.48\textwidth}{!}{
    \setlength{\tabcolsep}{1mm}{
    \begin{tabular}{l|cccc|c}
    \toprule[1.5pt]
 Officehome & {Domain-P}  & {Domain-A}   & {Domain-C}  & {Domain-R}  &Avg $\uparrow$ \\ 
 \midrule
HPS (ResNet)& 86.44 & 71.95 & 74.36 & 79.95 & 78.18\\
Router (ResNet)& 93.33 &77.05 &80.89 & 84.10 &83.84\\
Grad (ResNet)& 91.33 & 77.24 &80.89 & 86.64 &84.02\\
\midrule
HPS (Transformer)& 95.11 & 87.80 & 80.65 & 90.55 &88.53\\
Router (Transformer)& 95.78 & 89.02 & 81.82 & 89.63 & 89.06\\
Grad (Transformer)& 95.78 & 87.80 & 80.65 & 90.32 & 88.64\\
  \midrule
   \midrule
 VLCS & {Domain-V}  & {Domain-L}   & {Domain-C}  & {Domain-S}  &Avg $\uparrow$ \\ 
 \midrule
HPS (Resnet)& 85.59 &74.33 & 100.0 & 80.00 & 84.98\\
Router (Resnet)& 85.59 &79.14 & 100.0 & 80.87 &86.40\\
Grad (Resnet)& 87.29 & 74.87 & 100.0 & 83.48 &86.41\\
\hline
HPS (Transformer)& 91.10 & 78.07 & 100.0 & 84.78 & 88.49\\
Router (Transformer)& 92.37 & 79.68 & 100.0 & 85.22 & 89.32\\
Grad (Transformer)& 92.37 & 79.14 & 100.0 & 86.09 & 89.40\\
    \bottomrule[1.5pt]
    \end{tabular}}}
    \vspace{-0.4cm}
    \label{tab:transformer}
\end{table}

\begin{table}[t]
\caption{The impact of router sharing on performance. The experiment is based  on  Officehome and VLCS.}
    \centering
    \resizebox{0.48\textwidth}{!}{
    \setlength{\tabcolsep}{1mm}{
    \begin{tabular}{l|cccc|c}
    \toprule[1.5pt]
 Officehome & {Domain-P}  & {Domain-A}   & {Domain-C}  & {Domain-R}  &Avg $\uparrow$ \\ 
 \midrule
Shared (0+3) & 89.11 & 71.14 & 78.79 & 84.10 & 80.78 \\
Non-shared (0+3) & 87.56 & 64.63 & 71.79 & 77.65 & 75.41\\
 \midrule
Shared (2+3) & 90.22 & 68.29 & 78.79 & 82.95 & 80.06\\
Non-shared (2+3) & 89.33 & 70.33 & 79.72 & 80.65 & 80.01\\
  \midrule
   \midrule
 VLCS & {Domain-V}  & {Domain-L}   & {Domain-C}  & {Domain-S}  &Avg $\uparrow$ \\ 
 \midrule
Shared (2+3) & 84.75 & 72.73 & 100.0 & 81.30 & 84.69\\
Non-shared (2+3) & 82.20 & 73.80 & 100.0 & 79.57 & 83.89\\
 \midrule
Shared (0+1+2+3) & 83.47 & 77.54 & 100.0 & 80.00 & 85.25\\
Non-shared (0+1+2+3) & 83.90 & 73.80 & 100.0 & 80.00 & 84.42\\
\bottomrule[1.5pt]
    \end{tabular}}}
    \vspace{-0.4cm}
    \label{tab:routershare}
\end{table}

\par{\noindent \bf The impact of rank $r$.} 
We set different $r$ to evaluate the impact of the rank on performance (Tab.~\ref{tab:rank}). For both MoLA-Grad and MoLA-Router, a larger $r$ does not mean better performance. This may be due to over-fitting caused by the introduction of excessive parameters.

\par{\noindent \bf Comparison of using MoLA in different layer.}
Since our MoLA can be conveniently applied as a plug-in to any layer in the model, we compare the impact of applying MoLA at different positions in the model on performance.
From Tab.~\ref{tab:position}, we can observe that the performance of the model improves regardless of where MoLA is applied, demonstrating the effectiveness of MoLA in enhancing heterogeneous training. However, the highest performance is not achieved when MoLA is applied to all blocks. This suggests that different depth layers may exhibit varying degrees of heterogeneity, and blindly applying MoLA may lead to over-fitting and impact the learning of subsequent layers. We note that the best performance is obtained when MoLA is applied to Block4, indicating that the extraction of heterogeneous features primarily occurs in the deep layers.

\par{\noindent \bf The effectiveness of $\mathcal{L}_\mathrm{TwD}$.}
In MoLA-Router, we propose using $\mathcal{L}_\mathrm{TwD}$ to assist the router in learning different contribution weights for different tasks. To validate the effectiveness of $\mathcal{L}_\mathrm{TwD}$, we visualize the $\omega$ mentioned in Section~\ref{router}  in Fig.~\ref{fig:tsne}. 
It can be observed that without using $\mathcal{L}_\mathrm{TwD}$, the learned ${\omega}$ lacks discriminability. In this case, MoLA-Router fails to allocate different parameter combinations for different tasks.  However, after using $\mathcal{L}_\mathrm{TwD}$, ${\omega}$ exhibits significant differences,  demonstrating the effectiveness of our $\mathcal{L}_\mathrm{TwD}$.

\par{\noindent \bf Comparison of  different model structures.} 
{
We compare the improvements of MoLA under the ResNet and Transformer structures  in Table~\ref{tab:transformer}. Our observations can be summarized as: (1) Models based on the transformer architecture significantly outperform models based on the ResNet architecture in handling heterogeneous data. (2) When using the transformer architecture, our MoLA\_Grad and MoLA\_Router still outperform the baseline HPS, demonstrating the general effectiveness of introducing the MoLA structure for training heterogeneous data.
(3) MoLA\_Grad and MoLA\_Router show more significant improvements on small dataset (VLCS) for the transformer architecture.}

\par{\noindent \bf The impact of router sharing.} 
{
To evaluate the impact of router sharing, we conduct the experiments with independently configuring routers for different blocks, and the comparison results with using the shared router are shown in Table~\ref{tab:routershare}. Note that, the numbers in parentheses indicate which shared/unshared routers act on which residual blocks, and different block combinations are used to improve the generality of the conclusions.
Interestingly, we found that using shared routers can lead to better performance and fewer parameters than using non-shared routers. This is possibly because shared routers as a meta controller can capture more high-level information on allocating the parameter space by configuring experts, unlike the non-shared routers without any information exchange.}

\par{\noindent \bf The dimension of $\omega$ in MoLA\_Router.} 
{
We vary the feature dimension $d$ of $\omega$ to evaluate the hyper-parameter sensitivity  in Table~\ref{tab:omega}. We can see that it may not be necessary to improve the performance by increasing dimensions. The underlying reason is that such a setting may depend on the implicit heterogeneity of datasets. We empirically set $d$=32.}

\par{\noindent \bf The number of experts in MoLA\_Router.} 
{We vary the number of experts in MoLA\_Router and compare their performance in Table~\ref{tab:num_experts}.
It can be seen that increasing the number of experts may not absolutely bring the benefit to the training. Actually, this aligns the spirit of our design on primilary-secondary discrepancy. Namely, when we configure too many adapters, the common knowledge may not encompassed into the backbone, which reduces the benefit of heterogeneous training. However, when we configure too few adapters, we cannot mediate the conflicts. Therefore, this indeed depends on the heterogeneity of datasets. }

\begin{table}[t]
\caption{The impact of the feature dimension $d$ of $\omega$ in Router on performance. The experiment is based  on Officehome.}
    \centering
    \resizebox{0.48\textwidth}{!}{
    \setlength{\tabcolsep}{1mm}{
    \begin{tabular}{l|cccc|c}
    \toprule[1.5pt]
 Officehome & {Domain-P}  & {Domain-A}   & {Domain-C}  & {Domain-R}  &Avg $\uparrow$ \\ 
 \midrule
Router ($d$=32) & 90.67 & 74.39 & 80.42 & 82.95 & 82.11 \\
Router ($d$=64) & 82.89 & 68.29 & 72.73 & 75.58 & 74.87 \\
Router ($d$=128) & 90.00 & 69.11 & 78.79 & 82.03 & 79.98 \\
\bottomrule[1.5pt]
    \end{tabular}}}
    \vspace{-0.4cm}
    \label{tab:omega}
\end{table}

\begin{table}[t]
\caption{The impact of the number of experts ($n$) in Router  on performance. The experiment is based  on Officehome and VLCS.}
    \centering
    \resizebox{0.48\textwidth}{!}{
    \setlength{\tabcolsep}{1mm}{
    \begin{tabular}{l|cccc|c}
    \toprule[1.5pt]
 Officehome & {Domain-P}  & {Domain-A}   & {Domain-C}  & {Domain-R}  &Avg $\uparrow$ \\ 
 \midrule
Router ($n$=2) & 87.33 & 64.63 & 69.93 & 76.50 & 74.60 \\
Router ($n$=3) & 86.44 & 63.82 & 75.06 & 75.81 & 75.28\\
Router ($n$=4) & 89.78 & 73.17 & 80.19 &82.03 & 81.29\\
  \midrule
   \midrule
 VLCS & {Domain-V}  & {Domain-L}   & {Domain-C}  & {Domain-S}  &Avg $\uparrow$ \\ 
 \midrule
Router ($n$=2) & 85.17 & 74.87 & 100.0 & 81.74 & 85.44\\
Router ($n$=3) & 85.17 & 77.54 & 100.0 & 81.30 & 86.00\\
Router ($n$=4) & 85.17 & 72.73 & 99.00 & 79.13 & 84.01\\
\bottomrule[1.5pt]
    \end{tabular}}}
    \vspace{-0.4cm}
    \label{tab:num_experts}
\end{table}

\section{Conclusion}
This paper explores the issue of training conflicts in heterogeneous data training using Mixture of Low Rank Adapters (MoLA). Specifically, we propose two variants of MoLA, namely MoLA-Grad and MoLA-Router. The former uses task identifiers to assign specific low-rank adapters to each task, allowing task-specific gradients to be computed only within their respective adapters, thereby mitigating heterogeneity conflicts. The latter trains a router to adaptively assign different weight combinations of adapters for different tasks, achieving similar effects. We validate the effectiveness of MoLA in three common heterogeneous scenarios and present the in-depth analysis on its working mechanism.

\nocite{langley00}
\newpage
\section*{Acknowledgment} 
This work is supported by the National Key R\&D Program of China (No. 2022ZD0160702),  STCSM (No. 22511106101, No. 22511105700, No. 21DZ1100100), 111 plan (No. BP0719010) and National Natural Science Foundation of China (No. 62306178).

\section*{Impact Statement}
The method proposed in this paper can effectively improve the performance of heterogeneous data training and has a wide range of applications. 
Compared to previous methods, it significantly reduces the number of parameters and effectively saves computational costs. 
This exploration is of great significance to the directions that requires training with heterogeneous data, such as healthcare, general computer vision, and multimodal large models. 
In these fields, there is a large amount of heterogeneous training data, and there is a high demand for computational resources. 
So far, we have not discovered any negative impacts of this method.
{
\bibliography{main}
\bibliographystyle{icml2024}}

\newpage
\appendix
\onecolumn
\section{Datasets and Competing methods}

\textbf{Datasets.} \textbf{VLCS}~\citep{torralba2011unbiased} and \textbf{Officehome}~\citep{venkateswara2017deep} are well-known datasets for multi-domain  classification. VLCS encompasses 10,729 images across four domains: Caltech101 (C), LabelMe (L), SUN09 (S), and VOC2007 (V), organized into 5 classes. Office-Home comprises 15,588 images across  four domains: Art (A), Clipart (C), Product (P), and Real-World (R), spanning 65 classes. 
\textbf{RadImageNet}~\citep{mei2022radimagenet} and \textbf{MedMNISTV2}~\citep{yang2021medmnist} are typical medical image datasets that integrate different types of diseases from different parts of the body. RadImageNet is a large database with 1.35 million radiologic images from CT, MRI, and US modalities, annotated across 11 anatomical regions. MedMNISTV2 includes 18 types of medical data, and we select 9 types from it for heterogeneous training, totaling 340,803 samples.
\textbf{NYUv2}~\citep{silberman2012indoor} and \textbf{Cityscapes}~\citep{cordts2016cityscapes} are popular scene understanding datasets for MTL. 

\textbf{Competing methods.} We consider the following approaches in the filed of three data kinds with heterogeneity. In common, we apply several MTL methods to all kinds of data to reach fair comparison, including: (1) \textbf{Single-task}~\citep{sun2020adashare}: training a single network for each task individually; (2) \textbf{HSP}~\citep{wei2021hps}: introducing a multi-task network for medical image segmentation; (3) \textbf{Cross-Stitch}~\citep{misra2016cross}: representative soft parameter sharing based multi-task method; (4) \textbf{MMoE}~\citep{ma2018modeling}: introducing multi-gate to Mixture of adapters; (5) \textbf{CGC}~\citep{cheng2016cgc}: robust method for multi-domain graph clustering; (6) \textbf{LTB}~\citep{guo2020learning}: tree-structured multi-task learning network; (7) \textbf{DSelect-k}~\citep{hazimeh2021dselect}: improving MoE through continuous and sparse gate. In addition, we have also compared our method against approaches only suitable for specific scenarios, including domain heterogeneity methods: (1) \textbf{Uniform}: training a single network for all tasks uniformly; (2) \textbf{MMD-AAE}~\citep{li2018domain}: maximum mean discrepancy method for domain generalization; (3) \textbf{SelfReg}~\citep{kim2021selfreg}: a self-supervised contrastive regularization method for domain generalization; (4) \textbf{EQRM}~\citep{eastwood2022probable}: improving domain generalization through quantile risk minimization; (5) \textbf{DANN}~\citep{muralidhar2018incorporating}: incorporating prior domain knowledge into network, multi-input task heterogeneity methods: (1) \textbf{MGDA(-UB)}~\citep{sener2018multi}: multi-objective optimization method; (2) \textbf{IMTL}~\citep{liu2021towards}: impartial multi-task learning method; (3) \textbf{Nash-MTL}~\citep{navon2022multi}: view gradient combine for MTL as a bargaining game; (4) \textbf{MoCo}~\citep{he2020momentum}: momentum contrastive method for unsupervised visual representation learning  (5) \textbf{Aligned-MTL(-UB)}~\citep{senushkin2023independent}: improving MTL by independent component alignment, single-input task heterogeneity methods: (1) \textbf{RLW}: random loss weighting multi-task method; (2) \textbf{DWA}: dynamic weight averageing with attention; (3) \textbf{GradNorm}~\citep{chen2018gradnorm}: gradient normalization algorithm balancing multi-task;  (4) \textbf{GradDrop}: probabilistic masking procedure sampling gradients; (4) \textbf{PCGrad}~\citep{yu2020gradient}: gradient surgery for projection; (5) \textbf{GradVac}~\citep{wang2020gradient}: gradient vaccine encouraging  aligned parameter updates; (6) \textbf{CaGrad}~\citep{liu2021conflict}: conflict-averse gradient descent minimizing average loss. 

\section{Implementation Details}

We apply the AdamW optimizer~\citep{loshchilov2017decoupled} with learning rate of 0.0001 for experiments on VLCS and Officehome datasets, SGD optimizer~\citep{robbins1951stochastic} with learning rate of 0.05 on RadImageNet and MedMNIST datasets and Adam optimizer~\citep{kingma2014adam} with learning rate of 0.0001 on NYUv2 dataset. For all of the experiments, the training batch-size is set to 128. For experiments on VLCS, Officehome, RadImageNet and MedMNIST, we report the classification accuracy of separate domains and downstream tasks together with the average accuracy score. For NYUv2, we report mIoU and pixel-level accuracy (Pix Acc) for sementic segmentation, and average relative error (Rel.) for depth estimation and mean and average of angle error (Angle Dist.) and pixel accuracy as percentage of pixels with angle error below threshold $t$ for surface normal prediction. Additionally, we calculate the task-weighted scores to evaluate the overall performance on three tasks.

\end{document}